\documentclass[]{ceurart}

\usepackage{listings}
\usepackage{xcolor}

\usepackage{hyperref}
\usepackage[utf8]{inputenc}
\usepackage[T1]{fontenc}

\usepackage{graphicx}

\begin{document}

\copyrightyear{2025}
\copyrightclause{Copyright for this paper by its authors.
  Use permitted under Creative Commons License Attribution 4.0
  International (CC BY 4.0).}

 \conference{IberLEF 2025, September 2025, Zaragoza, Spain} 

\title{MRT at IberLEF-2025 PRESTA Task: Maximizing Recovery from Tables with Multiple Steps}

\tnotemark[1]


  


\author[1]{Maximiliano Hormaz\'abal Lagos}[%
orcid=0009-0003-3687-1924,
email=mhormazabal@gradiant.org,
]
\cormark[1]
\fnmark[1]
\address[1]{Fundaci\'on Centro Tecnol\'oxico de Telecomunicaci\'ons de Galicia (GRADIANT), Vigo, Spain}

\author[1]{\'Alvaro Bueno S\'aez}[%
orcid=0009-0006-9199-972X,
email=abueno@gradiant.org,
]
\cormark[1]
\fnmark[1]

\author[1]{H\'ector Cerezo-Costas}[%
orcid=0000-0003-2813-2462,
email=hcerezo@gradiant.org,
]
\cormark[1]
\fnmark[1]

\author[1]{Pedro Alonso Doval}[%
orcid=0009-0000-8255-3466,
email=palonso@gradiant.org,
]
\cormark[1]
\fnmark[1]

\author[1]{Jorge Alcalde Vesteiro}[%
orcid=0009-0003-6418-3694, 
email=jalcalde@gradiant.org,
]
\cormark[1]
\fnmark[1]
\cortext[1]{Corresponding author.}
\fntext[1]{These authors contributed equally.}

\begin{abstract}
This paper presents our approach for the IberLEF 2025 Task PRESTA: Preguntas y Respuestas sobre Tablas en Español (Questions and Answers about Tables in Spanish). 
Our solution obtains answers to the questions by implementing Python code generation with LLMs that is used to filter and process the table.  
This solution evolves from the MRT implementation for the Semeval 2025 related task. The process consists of multiple steps: analyzing and understanding the content of the table, selecting the useful columns, generating instructions in natural language, translating these instructions to code, running it, and handling potential errors or exceptions. These steps use open-source LLMs and fine-grained optimized prompts for each step. With this approach, we achieved an accuracy score of 85\% in the task.
\end{abstract}


\maketitle

\section{Introduction}

Natural Language Processing (NLP) is nowadays constrained by the amount of information that can be processed by Large Language Models (LLMs) due to the limited capacity of input data that  they can handle. Some applications that have this limitation are response generation using RAG systems in which the LLM generates answers with a small subset of retrieved documents\cite{liu-etal-2024-rethinking} as context. In table questions answering this limitation is aggravated as tables and databases can have millions of records and columns \cite{ruan2024languagemodelingtabulardata}, which by today standards will not fit completely in the LLM context.  

In this paper, we improve the algorithm of Maximizing Recovery from Tables with Multiple Steps (MRT)\cite{hormazabal-et-al}, a multi-step process that implements LLMs and Python code generation to answer questions as objectively as possible. Our system implements a sequential divide-and-conquer approach in which LLMs or heuristic algorithms are executed at each step with very specific tasks. These steps range from describing the tables and generating instructions in natural language to producing the source code to implement the previous instructions, executing it, and parsing the output to obtain the final answer. In comparison with end-to-end solutions, our approach is more explainable as it is easy to debug and trace which was the cause of a good or bad response. 

This paper addresses the  IberLEF 2025 \cite{iberlef2025overview} Task PRESTA: Preguntas y Respuestas sobre Tablas en Español (Questions and Answers about Tables in Spanish) \cite{iberlef2025prestaoverview}. The code that generated these results is publicly available\footnote{\url{https://github.com/Gradiant/MRT_TableQA/releases/tag/v2.0.0}}.

\section{Background}

As the name implies, question answering (QA) consists of answering questions that normally have objectively correct answers. Tabular QA requires the system to retrieve responses from knowledge bases represented as datasets in tables. Recent methods for QA in tabular data, such as TAPAS\cite{Herzig_2020}, TAPEX \cite{liu2021tapex} or Omnitab\cite{jiang2022omnitab}, integrate transformers with architectures specifically adapted to extract answers directly from tables as they integrate said tables as context. Parallel to this, LLMs have also been employed in zero-shot and few-shot strategies, and have proven to show high levels of usability in QA due to their prior knowledge. One of the benefits of zero/few shot strategies is the fact that domain-specific fine-tuning could be avoided, eliminating the usual needs of gathering, cleaning, and performing human validation for the task. In addition, recent LLMs include reasoning skills by default. This helps, but still presents difficulties with complex queries, involving multiple columns, large tables, or ambiguous questions requiring common-sense knowledge.  

Another approach is to parse natural language questions into formal queries in programming languages such as SQL. There are systems like Seq2SQL\cite{liu2024survey} or TableGPT2\cite{su2024tablegpt2}, which are designed to create SQL queries from relational database queries or Python code, respectively. These methods are theoretically independent of table size (there is no context limitation) and provide greater transparency by including intermediate steps to supervise generated queries. 

Several datasets to evaluate TableQA strategies have been released in the past years. WikiSQL \cite{zhong2017seq2sqlgeneratingstructuredqueries} based on Wikipedia and TabFact \cite{chen2020tabfactlargescaledatasettablebased} provided structured evaluation of tabular data. However, these datasets do not convey the heterogeneity of real-world tabular data, which are usually more complex, less homogeneous, and unstructured. \cite{hwang2019comprehensiveexplorationwikisqltableaware}. To address this problem, DataBench\cite{oses-grijalba-etal-2024-question} has been developed, which brings together $65$ real-world datasets with more than $1,300$ manually crafted question-answer pairs in multiple domains.

Apart from the methods with LLMs and code generators mentioned above, others use alternative strategies such as Retrieval Augmented Generation (RAG) or Chain-of-Thoughts (CoT). For example, TableRag  \cite{chen2024tableragmilliontokentableunderstanding} proposes the use of RAG systems for tabular comprehension tasks, such as QA, employing techniques such as query expansion and a double transformation to query languages. This process translates, on the one hand, the schema to be interacted with and, on the other hand, the operation necessary to identify the cells with the answer. A noteworthy proposal is Chain-of-Table \cite{wang2024chainoftableevolvingtablesreasoning}, which implements CoT as an iterative reasoning mechanism. Instead of executing the code in one shot, programming instructions are executed iteratively to add or discard information from the table until the final answer is found.

Although there has been recent progress in this field, certain challenges still persist, such as the enhancement of reasoning across multiple rows and columns, managing multiple domains and languages, mixing data from multiple tables, or improving explainability.


\section{Background of this task}

Similar tasks to PRESTA challenge have been recently published. This is the case of SemEval 2025 Task 8: Question-Answering over Tabular Data challenge  \cite{osesgrijalba-etal-2025-semeval-2025}, a task with the same purpose as this, mainly in English, with tables from other contexts and a wider range of topics.

For this task, we developed MRT: Maximizing Recovery from Tables with
Multiple Steps \cite{hormazabal-et-al}. 
This solution treats the tables as Pandas Dataframes and uses LLMs to generate Python code that could extract an answer to each question. We now present briefly the MRT system, which is our starting point in the PRESTA challenge. 

The initial step is the \textit{column descriptor} module, which analyses the data of each column (description of the column content meaning, type of data, frequent values, max and min values, etc.).
Secondly, the \textit{explainer} module, using the previous analysis, generates natural language instructions with an LLM. These instructions contain the steps needed to get the answer.
Then, the \textit{coder} module generates, also with an LLM, Python code from the text instructions, and afterward, this code is executed by the \textit{runner} module. If an exception occurs during the code execution or answer parsing, the system steps back
into the \textit{coder} in an iterative looping process until it gets a valid answer or a limit of attempts is exceeded. Finally, the \textit{interpreter} module and \textit{formatter} module implement different approaches for obtaining the answer in the desirable data type in order to match the expected result for the task. In each step, the same or different LLM could be used. Usually, we employ an LLM finetuned for code generation within the \textit{coder} module.

Some of the limitations of MRT were:
\begin{itemize}
    \item It struggles to filter categorical values when the value in the dataset has a different representation as it appears in the question. For example, when asked for \textit{Obama} the system may not find the value with a strict match if the representation in the table is \textit{Barack Obama} instead.
    \item The generated natural language instructions were more intricate than they need to be to get the right response. Sometimes filtering instructions were added that are not needed to obtain the response. The generated code produced several exceptions with filters that at first glance are actually easy to execute.
    \item It does not scale well with a large number of columns. Some modules include in the prompt all the columns of the table, their descriptions and statistical information and frequent values. However, this does not scale properly when the number of columns is large.
\end{itemize}

\section{System overview}

\begin{figure}[h]
    \centering
    \includegraphics[width=\textwidth]{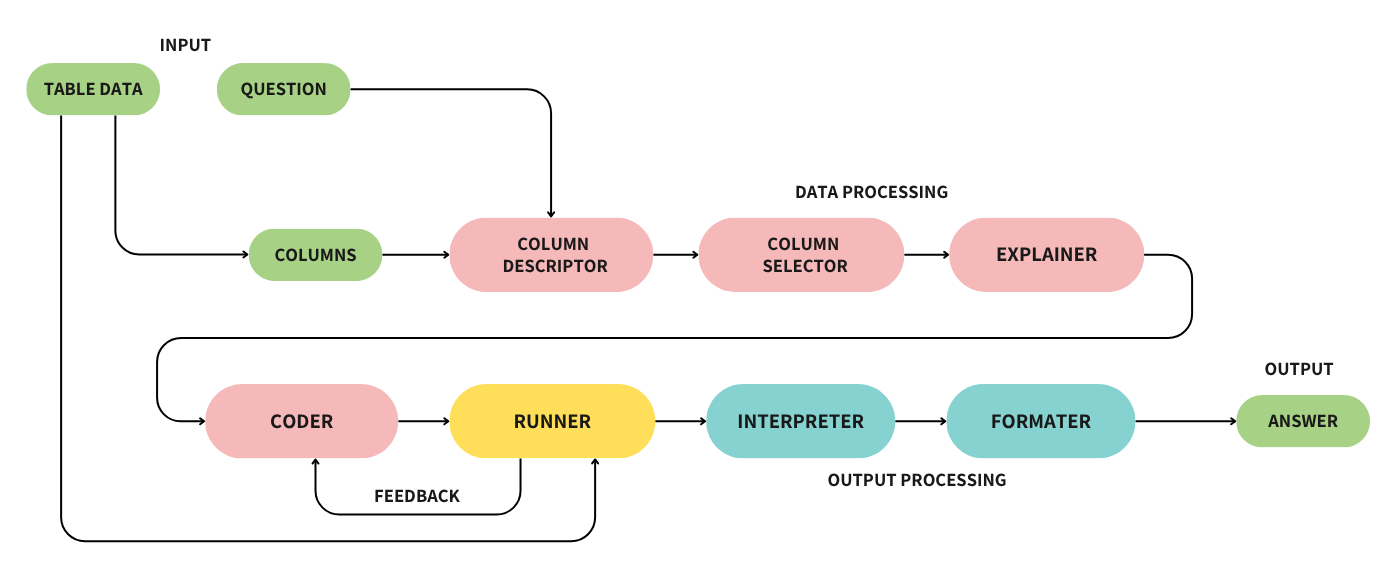}
    \caption{Diagram of the system showing all the steps involved in the generation of the response.}
    \label{fig:architecture}
\end{figure}

The system with the different components is presented in Figure \ref{fig:architecture}. New components were added from the previous version, \textit{column selector} and others were deeply modified (\textit{explainer}, \textit{coder} to make the system more resilient to the new challenges posed by the PRESTA challenge.

\subsection{New challenges addressed in this task}

Compared with the SemEval task, the PRESTA dataset patented new limitations that our former MRT system had:

\begin{itemize}
\item A larger number of columns in each table implied enormous prompts in the \textit{explainer} module. Sometimes, this leads to exceptions running LLMs. Having more columns increases the changes of selecting wrong columns to answer a question. For comparison, Semeval task tables had an average of $24.8$ columns, while in this IberLEF task they have an average of $174.1$ columns. 

\item Ambiguous names. Some columns have names that without all the context (or even with it) are not informative about the content of their cells, or that use initials that are not intuitive. In those cases, their values are difficult to interpret even by humans. Some clarifying examples are \textit{N\_R}, \textit{¿Usted es? (LEER\_S\'oLO PARA LOS QUE HAN CONTESTADO QUE "TRABAJA" EN LA P2014)} or other boolean columns that are possible answers to question, which could not be easily inferred by the name of the column: \textit{Analg\'esico\_antiinflamatorio\_1}, \textit{Antidepresivos (fluoxetina, sertralina, escitalopram)\_1}...

\item Columns with mixed types. For example, columns that are essentially numerical but actually contain strings for some of the values. This is the case of columns like \textit{Del 1 al 10, con qu\'e probabilidad votar\'ias al partido pol\'itico BNG?} (From 1 to 10, with which probability you would vote to BNG party?), whose values are $5$, $6$, $7$, etc., but also contain these options \textit{1 - No le votar\'ia nunca}, \textit{10 - Le votar\'ia siempre}. Using numerical operations directly in these columns would throw exceptions.

\end{itemize}

\subsection{Development of new features}

To mitigate and correct both the already known limitations inherited from the original MRT work and face the new challenges found in the training dataset for PRESTA, we developed and improved some features.

\subsubsection{Column selector module}

We designed the \textit{column selector} module to make an initial filter of columns before feeding the prompt of the \textit{explainer}, to avoid huge prompts that can lead to exceptions or errors. Due to its nature in the current implementation, the \textit{explainer} has to know in advance all the columns involved in answering the question and their descriptions in order to obtain correct natural language instructions. Those questions must be included in the prompt with the question. Henceforth, this new module tries to filter the columns to avoid crashes by large prompts whilst leaving the relevant questions untouched. \\

This new step uses an LLM to ask which columns are potentially relevant to that question. To achieve this, it prompts iteratively the LLM in groups of 25 columns each, giving their names and descriptions along the question. The prompt emphasizes that, in case of doubt, it should return the column, in order to not miss the relevant information in this step.  \\

The output of this module is the set of columns that the LLM considered useful for the \textit{explainer} module, conditioned by the query.

\subsubsection{Rules for removing uninformative columns}

Column names can be ambiguous and mislead the different modules. We identified exceptional columns that led to significant and recurrent errors, that we consider are not needed in any of the questions seen and will not make sense to use in questions of this style.
This is the case of \textit{"N\_R\_" ("No Recuerda"/"Don't remember")}, which the \textit{column descriptor} wrongly described as "Number of respondents". Having bad descriptions will produce wrong answers if those columns are used to obtain the answer instead of the good ones.

Columns consisting of numerations or that share almost the name with others, where it is difficult to identify semantic differences between them, were discarded. For example, there is one table that has columns identified as \textit{"Ns\_Nc\_0"}, \textit{ "Ns\_Nc\_1"}, \textit{"Ns\_Nc\_2"}, etc.

\subsubsection{Clarification instructions in natural language explanations}

The original MRT code returned raw text instructions that were then parsed to obtain a list of steps that should be coded afterwards.

The evolution over the original MRT implementation corrects mistakes due to the wrong naming of columns or variables in the natural language instructions.

In the new version, the raw text format using in the output was changed to a JSON with the following properties: \textit{instructions}: the list of natural language instructions that must be applied to the data to get the answer, \textit{columns}: the list of columns used in the instructions and \textit{filter\_values}, values that will be used to filter the information of the table.

After the natural language instructions are calculated a check is performed to correct the column names that are misspelled by the model. Levenshtein distance is used to obtain the closest column to the one appearing in the instructions, directly switching the name if they are not equal.

With the \textit{filter\_values} a similar procedure was implemented. Instead of substituting the values, a clarification instruction was added to those originally generated by the system including the old value. Those instructions have the following format \textit{"Be careful!. The value $\langle$old value$\rangle$ appears in the database with the following format: $\langle$new value$\rangle$"}.

Furthermore to simplify the task for the \textit{coder} information about the column types and the typical values of the column (only for the not-numerical columns) are added at the end of the instructions: \textit{"The column $\langle$column name$\rangle$ is of type $\langle$column type$\rangle$ and has the following example values: $\langle$column values$\rangle$"}.

Table \ref{tab:explainer_clarification} shows an example without clarification instructions and with clarification instructions. The instructions on the right are more complete and easier to understand than their counterparts on the left that lack certain information and are more prone to fail.

\begin{table}[h!]
\centering
\begin{tabular}{|l|l|}
\hline \textbf{W/o clarification Instructions} & \textbf{With clarification Instructions} \\
\hline
1) Count the total number of surveys conducted in & 1) Count the total number of surveys conducted in \\
January & January \\ 
2) Compare the count of surveys conducted in & 2) Compare the count of surveys conducted in \\
January with the total count of surveys to determine & January with the total count of surveys to determine \\
if most surveys were conducted in January. & if most surveys were conducted in January. \\
& 3) Be careful! The value enero appears in the \\
& database with the following format: 'Enero' \\
& 4) The column 'Mes de realizaci\'on' is of type 'object'  \\
& and has the following example values: Enero, \\
& Febrero, Marzo \\

\hline
\end{tabular}
\caption{Natural language instructions generated by the \textit{explainer} with and without clarification instructions for the question \texttt{¿Fueron la mayor\'ia de las encuestas realizadas en enero?}}
\label{tab:explainer_clarification}
\end{table}

\subsubsection{Custom Functions for Code Generation}

Assigning the full responsibility of both orchestrating and generating code to a single model could overload its capabilities. Empirical results supported this hypothesis, as evidenced by recurrent instances of poor coding practices despite explicit instructions to avoid them. Common problematic patterns included:

\begin{enumerate}
    \item Misuse of the \texttt{group\_by} function, leading to code errors instead of employing suitable alternatives that are less prone to produce exceptions.
    \item Contradictory ordering operations, such as sorting in descending order initially, followed by an explicit ascending sort, effectively negating previous instructions. 
\end{enumerate}

Inspired by the methodology of \cite{zhou2022tacubeprecomputingdatacubes}, though without precomputing function-result cubes, we developed instead pre-coded generic functions that can be used in multiple contexts but that solve many of these common mistakes. The prompt conditions the model to use these functions as an alternative to the \textit{pandas} implementations. This allowed the model to primarily focus on code orchestration, significantly reducing its code-generation workload.

The objective behind these functions was to maintain general applicability rather than targeting overly specific scenarios. For instance, functions are generalized to handle common data tasks, such as counting occurrences or filtering data based on numerical conditions.

These functions were created through a semi-automated approach involving both Large Language Models (LLMs) and human input:

\begin{enumerate}
    \item An LLM analyzed the training dataset and proposed generic function templates that could broadly address the questions of the dataset. Additionally, the LLM had the option to reuse existing templates. The outcome was a set of function templates capable of solving a substantial portion of the queries. This functions were obtained from the training split.
    \item Human developers implemented these function templates, subsequently verifying that the model actively employed them, therefore improving empirical performance.
    \item During the testing and debugging phase, we validated that pre-coded, generalized functions effectively resolved recurring issues, prompting further expansion of the function pool.
    \item Additionally, certain functions incorporated fuzzy decision-making capabilities to enhance performance, a methodology detailed in the subsequent section, \textit{Fuzzy Search of Categorical Values}.
\end{enumerate}

The final set of developed generic functions is in appendix I.

Additionally, we specifically addressed the challenge of columns misidentified as numerical due to naming conventions through the dedicated internal function \texttt{extract\_numeric}, detailed in the Appendix \ref{AppendixI}.

Lastly, another way to conceptualize this methodology is by viewing the provision of these functions as enabling the model to utilize Python tool-calling capabilities, directly resolving common coding problems encountered by the model \cite{he2024achievingtoolcallingfunctionality}.

\subsubsection{Fuzzy Search of Categorical Values}

As previously described, certain functions within the model pipeline employ fuzzy matching techniques to mitigate typical errors encountered during data processing, particularly those stemming from inconsistencies or typographical variations in categorical values. These errors frequently arise due to the complexity and variability of data names, especially after undergoing multiple processing stages.

To clarify the motivation behind implementing fuzzy matching, we first outline the challenges experienced in earlier iterations of the pipeline and also in the version presented in this paper. Specifically, accessing column and row names in a reliable manner posed substantial difficulties due to discrepancies that emerged across the five stages of large language model (LLM) processing. Such discrepancies typically result in false positives and false negatives:

\begin{enumerate}
    \item \textbf{False positives} occur when the model incorrectly forwards values due to minor deviations in the text, such as pluralization or capitalization inconsistencies. For example, the value "item" might erroneously be transformed into "items" or "Item."
    \item \textbf{False negatives} occur when originally correct but unusually formatted values are improperly corrected, thereby introducing errors. For instance, the value "iTem" might incorrectly be normalized to "item," altering the intended representation.
\end{enumerate}

To address these challenges, we adopt established fuzzy matching methodologies as detailed in prior research such as \cite{1423558}. Rather than relying solely on exact matches—which fail to accommodate minor textual variations—fuzzy matching techniques allow the model to recognize and utilize values that closely resemble the target values based on a defined similarity threshold.

An illustrative example of such a fuzzy matching approach in our pipeline is presented through the Python functions shown in appendix \ref{AppendixII}.

These functions perform sequentially the following steps:

\begin{enumerate}
    \item Identifying the target column and value.
    \item Recognize empirically that exact matches are often not achievable due to textual variations.
    \item If an exact match is unavailable, apply fuzzy matching in a secondary step, selecting the closest matching value based on the predefined similarity threshold and subsequently operating on it.
\end{enumerate}

In practice, this fuzzy matching strategy enhances significantly the robustness and overall accuracy of the pipeline, effectively reducing the incidence of errors due to minor textual discrepancies.



\section{Experimental setup}
The experimental setup was executed in batches, where all the questions in the dataset were run through one step before advancing to the following one. Thus, the number of times a model has to be loaded/unloaded was optimized as each of the steps may use different models. Also, results of the column descriptor were cached between experiments and was executed only once, given that its output for a table is independent of the questions.

The tests were executed in a NVIDIA RTX-a6000 that combines 84 second-generation RT cores, 336 third-generation Tensor cores, and 10,752 CUDA cores with 48 GB of graphics memory for performance.

\subsection{Dataset splits}

Although no training of any model has been performed, the splits of the dataset are shown below (see table \ref{tab:dataset_splits}). The tables used in the test consist of the same tables of the train and the dev splits.

\begin{table}[h!]
\centering
\begin{tabular}{|c|c|c|}
\hline
\textbf{Split} & \textbf{Tables} & \textbf{Questions} \\
\hline
train & 6 & 150 \\
\hline
dev & 4 & 100 \\
\hline
test & 10 & 100 \\
\hline
\end{tabular}
\caption{Distribution of number of tables and questions for each split in the dataset}
\label{tab:dataset_splits}
\end{table}

Train and dev splits have been used for the development of the modules, whereas test split was solely used for the validation of the system against the official platform used in the benchmark.

\subsection{Models} \label{subsec:models}
We decided to use Qwen\footnote{https://huggingface.co/Qwen} models for the different modules of the system. 
Specifically, we executed two types of Qwen models: Qwen 2.5 14B\footnote{https://huggingface.co/Qwen/Qwen2.5-14B-Instruct} for all the modules, excepting the coder which used Qwen 2.5 coder 14B\footnote{https://huggingface.co/Qwen/Qwen2.5-Coder-14B-Instruct}.

We also made some tests using the recently published model Qwen 3 \footnote{https://huggingface.co/Qwen/Qwen3-14B} in the explainer, maintaining Qwen 2.5 and Qwen 2.5 coder for the rest of the modules.

\section{Results}

\subsection{Performance in validation test}
Table \ref{tab:scores_dev} shows the results obtained in validation data in our main scenario (executing with Qwen 2.5 and Qwen 2.5 coder). We can see the performance for each type of data. The total score is 71\%, while for most of the data types are very homogeneous, varying from 75\% to 80\%, except for numerical answers, which clearly perform worse with a 50\% score.

\begin{table}[h!]
\centering
\begin{tabular}{|c|c|c|c|c|c|c|}
\hline \textbf{ } & \textbf{Total} & \textbf{Boolean} &
\textbf{Number} & \textbf{Category} & \textbf{List[Category]} & \textbf{List[Number]}\\
\hline
 Score & 0.71 & 0.75 & 0.5 & 0.77 & 0.8 & 0.75\\
\hline
Size & 100  & 20 &22  & 22 &  20 & 16 \\
\hline
\end{tabular}
\caption{Score and number of answers per answer type in the validation split.}
\label{tab:scores_dev}
\end{table}

\subsection{Performance in test set}
We submitted 3 results to the task. One is with the results of our system using Qwen 2.5 14B for all the steps that involve LLMs, except the coder which used Qwen Coder 14B. The second one selects the output of the interpreter module in that same execution. Finally, a third one changes only the model of the explainer to Qwen 3 14B. They achieve \textbf{scores of 85\%, 85\%, and 83\%} respectively. We show in the table \ref{tab:scores_test} our best submission broken down by type of expected answer. All the experiments were repeated $8$ times, taking the most repeated answer (a simple majority voting strategy) as the final result.


\begin{table}[h!]
\centering
\begin{tabular}{|c|c|c|c|c|c|c|}
\hline \textbf{ } & \textbf{Total} & \textbf{Boolean} &
\textbf{Number} & \textbf{Category} & \textbf{List[Category]} & \textbf{List[Number]}\\
\hline
 Score & 0.85 & 0.95 & 0.9 & 0.8 & 0.8 & 0.75\\
\hline
Size & 100  & 20 & 20  & 20 &  20 & 20 \\
\hline
\end{tabular}
\caption{Score and number of answer per answer type in the test split.}
\label{tab:scores_test}
\end{table}

In comparison with the score of validation set, in this case is 15\% higher. It matches our intuition that is that the test set questions are in average simpler than the validation set questions.

\subsection{Manual Error Analysis in the validation set}
We performed a manual error analysis of the answers for the validation set flagged as an error by the evaluator. The results are summarized in Table \ref{table:manual_error_analysis}.
The main source of errors is the wrong generation of instructions in the \textit{explainer} module. Some of the errors involve not selecting the correct column to use, although sometimes the ambiguity of the column names makes this choice difficult. For example, columns such us 'Edad' vs 'Edad\_recodificada' in which one is just a higher level of abstraction from the other. Other cases involved just wrong natural language instructions. Adding unnecessary extra filters (removing nulls, zeros, empty lists, etc.) was very frequent source of errors.

The other main source of errors is the removal of relevant columns in the column selector. Sometimes it struggles with columns that have long names and even involve complex semantics such as conditionals, like survey questions present in this dataset.

We identified a few errors in the interpreter for cases where the runner actually obtained the correct response. This case usually involves incorrectly removing symbols or clarifications. For instance, there are two related to age intervals: "+65" and "18-24" which were changed to "65" and "1824". Other less frequent errors were due to the transformations of the output that make the metric implemented to consider it as an error. The removal of the parenthesis and the information within in 'PP' in  the expected answer 'PP (Partido Popular)'.
This last example is flagged as incorrect according to the benchmark metrics but that could be perfectly be deemed as acceptable by common sense with human feedback. 

Compared to manual error analysis in our previous work for the analogous SemEval task, we can prove the benefits of some of our new features taking into account that wrong cell filtering is not an issue anymore as it was before and code errors and code exceptions have been notably reduced.


\begin{table}[h!]
\centering
\begin{tabular}{|c|c|c|}
\hline
\textbf{Description} & Errors &\textbf{\% error} \\
\hline
Wrong Instructions & 11 & 37.9\%\\ 
\hline
Wrong column filtering & 9 & 31.0\%\\
\hline
Formatting (transformations) & 3 & 10.3\% \\
\hline
Code Generation  (incl. exceptions) & 2 & 6.6\%\\
\hline
Others & 4 & 13.7\% \\ 
\hline
\end{tabular}
\caption{Manual analysis of the errors in the validation split}
\label{table:manual_error_analysis}
\end{table}

\subsection{Ablation study}
We have performed an ablation study to evaluate the impact of some of the features on the performance of our system. To do this, we deactivate one by one the new modules and execute the benchmark with the validation set. Every configuration is repeated $8$ times, taking a majority voting ensemble as the final result, discarding thrown exceptions or bad results (e.g. responses such as \textit{No matching records were found}).

Table \ref{tab:ablation_study} shows the scores of each of the tries broken down by the type of expected answer. As can be seen, the overall metric is always in the range of $0.69$ and $0.74$. The number of question-answer pairs is very low ($100$) and the diversity of tables used in the validation set ($4$) is not enough to confirm whether the new modules improved or not the performance of the system in the test. For example, the \textit{column selector} module was implemented to avoid throwing exceptions when the number of columns was very large. Nevertheless, none of the $4$ datasets used in the validation throws this exception. Hence filtering the columns could have a detrimental effect if a good column was removed. Filtering out columns has a positive impact on time consumption, as the execution of the overall system is much faster (e.g. about three times in our experiments within the PRESTA dataset).

\begin{table}[h!]
\centering
\begin{tabular}{|c|c|c|c|c|c|c|}
\hline
\textbf{Scenario} & \textbf{Score in dev} & \textbf{Boolean} &
\textbf{Number} & \textbf{Category} & \textbf{List} & \textbf{List} \\
\textbf{Scenario} & &  & & & \textbf{Category} & \textbf{Number} \\
\hline
All (formatter) & 0.71 & 0.75 & 0.5 & 0.77 & 0.8 &	0.75\\
\hline
All (Interpreter) & 0.71 & 0.75 & 0.5 & 0.77 & 0.8 &	0.75\\
\hline
w/o column selector & 0.74 & 0.8 & 0.55 & 0.82 & 0.8 &	0.75\\
\hline
w/o custom functions & 0.72 & 0.75	& 0.55 & 0.77 & 0.75 & 0.81 \\
\hline
w/o explainer corrector & 0.7 & 0.7  & 0.5 & 0.77	& 0.8 & 0.75 \\
\hline
w/o retries in coder & 0.71	& 0.7	& 0.5	& 0.77	& 0.85	& 0.75	\\
\hline
w/o fuzzy subs. & 0.69	& 0.8	& 0.5	& 0.73	& 0.8	& 0.63	\\
\hline

\hline
\end{tabular}
\caption{Score in each of the scenarios in the validation set of the ablation study broke down by type of data.}
\label{tab:ablation_study}
\end{table}

\begin{figure}[h]
    \centering
    \includegraphics[width=\textwidth]{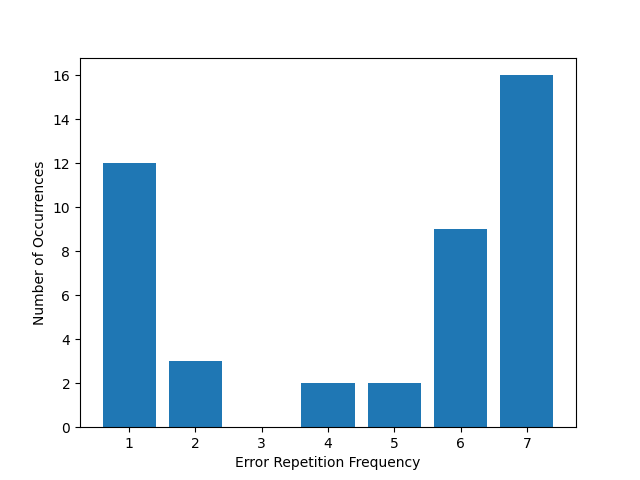}
    \caption{Error repetition frequency in all the $7$ experiments in the validation set}
    \label{fig:errorfreq}
\end{figure}

By seeing these results, one question that might arise is whether all the experiments are failing in the same questions. In other words, if the errors in the validation set are not addressed by the new modules. Figure \ref{fig:errorfreq} shows the error repetition frequency of errors in all the experiments. We can see that more than half of the errors are coincident in all the experiments. 

Combining the information of the most repeated errors ($6$ and $7$ times) with the information of the manual analysis, we find that most errors came from the bad selection of columns (9) or the \textit{explainer} not being able to generate good instructions in natural language (7). Some errors were formatting issues (2) and errors due to the validation metric but they were essentially correct (3).

\begin{figure}[h]
    \centering
    \includegraphics[width=\textwidth]{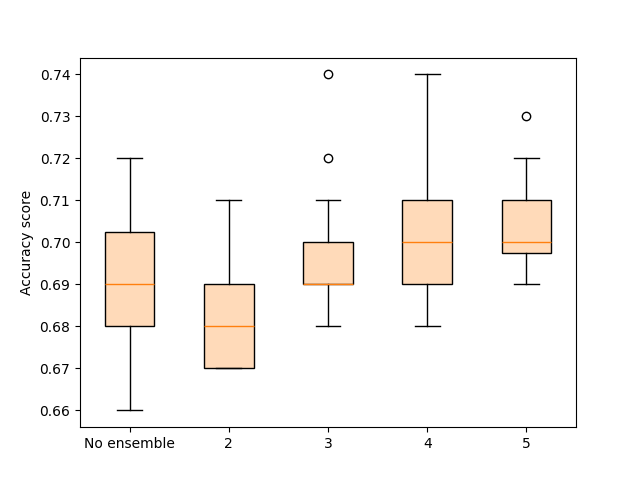}
    \caption{Accuracy score using an ensemble with majority voting}
    \label{fig:ensemble}
\end{figure}

Finally, we can observe the effect of using an ensemble with majority voting in Figure \ref{fig:ensemble}. An ensemble of $5$ experiments should be enough, although we have used $8$ repetitions in our configuration.

\section{Conclusions}

In this work, we presented our system MRT for answering queries over tables. Our system generates Python code involving multiple steps: describing and filtering columns, generating natural language instructions, code generation, and formatting the answer. This strategy builds upon our previous work in which we addressed some of the common sources of problems made by the previous version: adding auxiliary functions to avoid recurring exceptions, fuzzy match of cell and column names to improve the understanding of the question and instructions, and the selection of relevant columns to avoid LLM exceptions due to context size and improving the speed of the system at the same time. We used middle-size pretrained LMs with $14B$ parameters achieving a third place in the task with a $0.85\%$ of accuracy. One clear benefit of our approach is its explainability as the user can very easily understand what is the source of the errors by seeing the natural language instructions and the generated code.

However, evaluating the benefits of the theoretical improvements is difficult as the dataset lacks the size and diversity in order to be statistically relevant. The differences between the configurations are very small (between 1 and 5 question/answer pairs). In the future, we plan to test the system against larger datasets in order to gain more insights into the relevance of each block in the final answer.


\newpage
\section*{Declaration on Generative AI}
 During the preparation of this work, the authors used Grammarly in order to: Grammar and spelling check and GPT-4 in order to: suggestions for academic writing style. After using these tools, the authors reviewed and edited the content as needed and take full responsibility for the publication’s content. 

\newpage
\bibliography{sample-ceur}

\begin{thebibliography}{20}
\expandafter\ifx\csname natexlab\endcsname\relax\def\natexlab#1{#1}\fi
\providecommand{\url}[1]{\texttt{#1}}
\providecommand{\href}[2]{#2}
\providecommand{\path}[1]{#1}
\providecommand{\DOIprefix}{doi:}
\providecommand{\ArXivprefix}{arXiv:}
\providecommand{\URLprefix}{URL: }
\providecommand{\Pubmedprefix}{pmid:}
\providecommand{\doi}[1]{\href{http://dx.doi.org/#1}{\path{#1}}}
\providecommand{\Pubmed}[1]{\href{pmid:#1}{\path{#1}}}
\providecommand{\bibinfo}[2]{#2}
\ifx\xfnm\relax \def\xfnm[#1]{\unskip,\space#1}\fi
\bibitem[{Liu et~al.(2024)Liu, Wang, and Chen}]{liu-etal-2024-rethinking}
\bibinfo{author}{T.~Liu}, \bibinfo{author}{F.~Wang}, \bibinfo{author}{M.~Chen},
\newblock \bibinfo{title}{Rethinking {T}abular {D}ata {U}nderstanding with {L}arge {L}anguage {M}odels},
\newblock in: \bibinfo{editor}{K.~Duh}, \bibinfo{editor}{H.~Gomez}, \bibinfo{editor}{S.~Bethard} (Eds.), \bibinfo{booktitle}{Proceedings of the 2024 Conference of the North American Chapter of the Association for Computational Linguistics: Human Language Technologies (Volume 1: Long Papers)}, \bibinfo{publisher}{Association for Computational Linguistics}, \bibinfo{address}{Mexico City, Mexico}, \bibinfo{year}{2024}, pp. \bibinfo{pages}{450--482}. \URLprefix \url{https://aclanthology.org/2024.naacl-long.26/}. \DOIprefix\doi{10.18653/v1/2024.naacl-long.26}.
\bibitem[{Ruan et~al.(2024)Ruan, Lan, Ma, Dong, He, and Feng}]{ruan2024languagemodelingtabulardata}
\bibinfo{author}{Y.~Ruan}, \bibinfo{author}{X.~Lan}, \bibinfo{author}{J.~Ma}, \bibinfo{author}{Y.~Dong}, \bibinfo{author}{K.~He}, \bibinfo{author}{M.~Feng}, \bibinfo{title}{Language modeling on tabular data: A survey of foundations, techniques and evolution}, \bibinfo{year}{2024}. \URLprefix \url{https://arxiv.org/abs/2408.10548}. \href{http://arxiv.org/abs/2408.10548}{{\tt arXiv:2408.10548}}.
\bibitem[{Hormazabal-Lagos et~al.(2025)Hormazabal-Lagos, Saez, Cerezo-Costas, Doval, and Vesteiro}]{hormazabal-et-al}
\bibinfo{author}{M.~Hormazabal-Lagos}, \bibinfo{author}{A.~B. Saez}, \bibinfo{author}{H.~Cerezo-Costas}, \bibinfo{author}{P.~A. Doval}, \bibinfo{author}{J.~A. Vesteiro},
\newblock \bibinfo{title}{{MRT} at {S}em{E}val-2025 {T}ask 8: {M}aximizing {R}ecovery from {T}ables with {M}ultiple {S}teps},
\newblock in: \bibinfo{booktitle}{Proceedings of the 19th International Workshop on Semantic Evaluation (SemEval-2025)}, \bibinfo{publisher}{Association for Computational Linguistics}, \bibinfo{address}{Vienna, Austria}, \bibinfo{year}{2025}.
\bibitem[{Gonz\'alez-Barba et~al.(2025)Gonz\'alez-Barba, Chiruzzo, and Jim\'enez-Zafra}]{iberlef2025overview}
\bibinfo{author}{J.~A. Gonz\'alez-Barba}, \bibinfo{author}{L.~Chiruzzo}, \bibinfo{author}{S.~M. Jim\'enez-Zafra},
\newblock \bibinfo{title}{{Overview of IberLEF 2025: Natural Language Processing Challenges for Spanish and other Iberian Languages}},
\newblock in: \bibinfo{booktitle}{Proceedings of the Iberian Languages Evaluation Forum (IberLEF 2025), co-located with the 41st Conference of the Spanish Society for Natural Language Processing (SEPLN 2025), CEUR-WS. org}, \bibinfo{year}{2025}.
\bibitem[{Os\'es-Grijalba et~al.(2025)Os\'es-Grijalba, Ureña-L\'opez, C\'amara, and Camacho-Collados}]{iberlef2025prestaoverview}
\bibinfo{author}{J.~Os\'es-Grijalba}, \bibinfo{author}{L.~A. Ureña-L\'opez}, \bibinfo{author}{E.~M. C\'amara}, \bibinfo{author}{J.~Camacho-Collados},
\newblock \bibinfo{title}{{Overview of PRESTA at IberLEF 2025: Question Answering Over Tabular Data In Spanish}},
\newblock in: \bibinfo{booktitle}{Proceedings of the Iberian Languages Evaluation Forum (IberLEF 2025), co-located with the 41st Conference of the Spanish Society for Natural Language Processing (SEPLN 2025), CEUR-WS. org}, \bibinfo{year}{2025}.
\bibitem[{Herzig et~al.(2020)Herzig, Nowak, Müller, Piccinno, and Eisenschlos}]{Herzig_2020}
\bibinfo{author}{J.~Herzig}, \bibinfo{author}{P.~K. Nowak}, \bibinfo{author}{T.~Müller}, \bibinfo{author}{F.~Piccinno}, \bibinfo{author}{J.~Eisenschlos},
\newblock \bibinfo{title}{Ta{P}as: {W}eakly {S}upervised {T}able {P}arsing via {P}re-training},
\newblock in: \bibinfo{booktitle}{Proceedings of the 58th Annual Meeting of the Association for Computational Linguistics}, \bibinfo{publisher}{Association for Computational Linguistics}, \bibinfo{year}{2020}. \URLprefix \url{http://dx.doi.org/10.18653/v1/2020.acl-main.398}. \DOIprefix\doi{10.18653/v1/2020.acl-main.398}.
\bibitem[{Liu et~al.(2021)Liu, Chen, Guo, Ziyadi, Lin, Chen, and Lou}]{liu2021tapex}
\bibinfo{author}{Q.~Liu}, \bibinfo{author}{B.~Chen}, \bibinfo{author}{J.~Guo}, \bibinfo{author}{M.~Ziyadi}, \bibinfo{author}{Z.~Lin}, \bibinfo{author}{W.~Chen}, \bibinfo{author}{J.-G. Lou},
\newblock \bibinfo{title}{Tapex: Table {P}re-{T}raining via {L}earning a {N}eural {SQL} {E}xecutor},
\newblock \bibinfo{journal}{arXiv preprint arXiv:2107.07653}  (\bibinfo{year}{2021}).
\bibitem[{Jiang et~al.(2022)Jiang, Mao, He, Neubig, and Chen}]{jiang2022omnitab}
\bibinfo{author}{Z.~Jiang}, \bibinfo{author}{Y.~Mao}, \bibinfo{author}{P.~He}, \bibinfo{author}{G.~Neubig}, \bibinfo{author}{W.~Chen},
\newblock \bibinfo{title}{Omni{T}ab: {P}retraining with {N}atural and {S}ynthetic {D}ata for {F}ew-{S}hot {T}able-based {Q}uestion {A}nswering},
\newblock \bibinfo{journal}{arXiv preprint arXiv:2207.03637}  (\bibinfo{year}{2022}).
\bibitem[{Liu et~al.(2024)Liu, Shen, Li, Ma, Jiang, Zhang, Fan, Li, Tang, and Luo}]{liu2024survey}
\bibinfo{author}{X.~Liu}, \bibinfo{author}{S.~Shen}, \bibinfo{author}{B.~Li}, \bibinfo{author}{P.~Ma}, \bibinfo{author}{R.~Jiang}, \bibinfo{author}{Y.~Zhang}, \bibinfo{author}{J.~Fan}, \bibinfo{author}{G.~Li}, \bibinfo{author}{N.~Tang}, \bibinfo{author}{Y.~Luo},
\newblock \bibinfo{title}{A {S}urvey of {NL2SQL} with {L}arge {L}anguage {M}odels: {W}here are we, and {W}here are we {G}oing?},
\newblock \bibinfo{journal}{arXiv preprint arXiv:2408.05109}  (\bibinfo{year}{2024}).
\bibitem[{Su et~al.(2024)Su, Wang, Ye, Zhou, Zhang, Chen, Zhu, Wang, Xu, Chen et~al.}]{su2024tablegpt2}
\bibinfo{author}{A.~Su}, \bibinfo{author}{A.~Wang}, \bibinfo{author}{C.~Ye}, \bibinfo{author}{C.~Zhou}, \bibinfo{author}{G.~Zhang}, \bibinfo{author}{G.~Chen}, \bibinfo{author}{G.~Zhu}, \bibinfo{author}{H.~Wang}, \bibinfo{author}{H.~Xu}, \bibinfo{author}{H.~Chen}, et~al.,
\newblock \bibinfo{title}{Table{GPT2}: A {L}arge {M}ultimodal {M}odel with {T}abular {D}ata {I}ntegration},
\newblock \bibinfo{journal}{arXiv preprint arXiv:2411.02059}  (\bibinfo{year}{2024}).
\bibitem[{Zhong et~al.(2017)Zhong, Xiong, and Socher}]{zhong2017seq2sqlgeneratingstructuredqueries}
\bibinfo{author}{V.~Zhong}, \bibinfo{author}{C.~Xiong}, \bibinfo{author}{R.~Socher}, \bibinfo{title}{Seq2{SQL}: {G}enerating {S}tructured {Q}ueries from {N}atural {L}anguage using {R}einforcement {L}earning}, \bibinfo{year}{2017}. \URLprefix \url{https://arxiv.org/abs/1709.00103}. \href{http://arxiv.org/abs/1709.00103}{{\tt arXiv:1709.00103}}.
\bibitem[{Chen et~al.(2020)Chen, Wang, Chen, Zhang, Wang, Li, Zhou, and Wang}]{chen2020tabfactlargescaledatasettablebased}
\bibinfo{author}{W.~Chen}, \bibinfo{author}{H.~Wang}, \bibinfo{author}{J.~Chen}, \bibinfo{author}{Y.~Zhang}, \bibinfo{author}{H.~Wang}, \bibinfo{author}{S.~Li}, \bibinfo{author}{X.~Zhou}, \bibinfo{author}{W.~Y. Wang}, \bibinfo{title}{Tab{F}act: A {L}arge-scale {D}ataset for {T}able-based {F}act {V}erification}, \bibinfo{year}{2020}. \URLprefix \url{https://arxiv.org/abs/1909.02164}. \href{http://arxiv.org/abs/1909.02164}{{\tt arXiv:1909.02164}}.
\bibitem[{Hwang et~al.(2019)Hwang, Yim, Park, and Seo}]{hwang2019comprehensiveexplorationwikisqltableaware}
\bibinfo{author}{W.~Hwang}, \bibinfo{author}{J.~Yim}, \bibinfo{author}{S.~Park}, \bibinfo{author}{M.~Seo}, \bibinfo{title}{A {C}omprehensive {E}xploration on {W}iki{SQL} with {T}able-{A}ware {W}ord {C}ontextualization}, \bibinfo{year}{2019}. \URLprefix \url{https://arxiv.org/abs/1902.01069}. \href{http://arxiv.org/abs/1902.01069}{{\tt arXiv:1902.01069}}.
\bibitem[{Os\'es~Grijalba et~al.(2024)Os\'es~Grijalba, Ureña-L\'opez, Mart\'inez~C\'amara, and Camacho-Collados}]{oses-grijalba-etal-2024-question}
\bibinfo{author}{J.~Os\'es~Grijalba}, \bibinfo{author}{L.~A. Ureña-L\'opez}, \bibinfo{author}{E.~Mart\'inez~C\'amara}, \bibinfo{author}{J.~Camacho-Collados},
\newblock \bibinfo{title}{Question {A}nswering over {T}abular {D}ata with {D}ata{B}ench: A {L}arge-{S}cale {E}mpirical {E}valuation of {LLM}s},
\newblock in: \bibinfo{editor}{N.~Calzolari}, \bibinfo{editor}{M.-Y. Kan}, \bibinfo{editor}{V.~Hoste}, \bibinfo{editor}{A.~Lenci}, \bibinfo{editor}{S.~Sakti}, \bibinfo{editor}{N.~Xue} (Eds.), \bibinfo{booktitle}{Proceedings of the 2024 Joint International Conference on Computational Linguistics, Language Resources and Evaluation (LREC-COLING 2024)}, \bibinfo{publisher}{ELRA and ICCL}, \bibinfo{address}{Torino, Italia}, \bibinfo{year}{2024}, pp. \bibinfo{pages}{13471--13488}. \URLprefix \url{https://aclanthology.org/2024.lrec-main.1179/}.
\bibitem[{Chen et~al.(2024)Chen, Miculicich, Eisenschlos, Wang, Wang, Chen, Fujii, Lin, Lee, and Pfister}]{chen2024tableragmilliontokentableunderstanding}
\bibinfo{author}{S.-A. Chen}, \bibinfo{author}{L.~Miculicich}, \bibinfo{author}{J.~M. Eisenschlos}, \bibinfo{author}{Z.~Wang}, \bibinfo{author}{Z.~Wang}, \bibinfo{author}{Y.~Chen}, \bibinfo{author}{Y.~Fujii}, \bibinfo{author}{H.-T. Lin}, \bibinfo{author}{C.-Y. Lee}, \bibinfo{author}{T.~Pfister}, \bibinfo{title}{Table{RAG}: {M}illion-{T}oken {T}able {U}nderstanding with {L}anguage {M}odels}, \bibinfo{year}{2024}. \URLprefix \url{https://arxiv.org/abs/2410.04739}. \href{http://arxiv.org/abs/2410.04739}{{\tt arXiv:2410.04739}}.
\bibitem[{Wang et~al.(2024)Wang, Zhang, Li, Eisenschlos, Perot, Wang, Miculicich, Fujii, Shang, Lee, and Pfister}]{wang2024chainoftableevolvingtablesreasoning}
\bibinfo{author}{Z.~Wang}, \bibinfo{author}{H.~Zhang}, \bibinfo{author}{C.-L. Li}, \bibinfo{author}{J.~M. Eisenschlos}, \bibinfo{author}{V.~Perot}, \bibinfo{author}{Z.~Wang}, \bibinfo{author}{L.~Miculicich}, \bibinfo{author}{Y.~Fujii}, \bibinfo{author}{J.~Shang}, \bibinfo{author}{C.-Y. Lee}, \bibinfo{author}{T.~Pfister}, \bibinfo{title}{Chain-of-{T}able: {E}volving {T}ables in the {R}easoning {C}hain for {T}able {U}nderstanding}, \bibinfo{year}{2024}. \URLprefix \url{https://arxiv.org/abs/2401.04398}. \href{http://arxiv.org/abs/2401.04398}{{\tt arXiv:2401.04398}}.
\bibitem[{Os\'es~Grijalba et~al.(2025)Os\'es~Grijalba, Ureña-L\'opez, Mart\'inez~C\'amara, and Camacho-Collados}]{osesgrijalba-etal-2025-semeval-2025}
\bibinfo{author}{J.~Os\'es~Grijalba}, \bibinfo{author}{L.~A. Ureña-L\'opez}, \bibinfo{author}{E.~Mart\'inez~C\'amara}, \bibinfo{author}{J.~Camacho-Collados},
\newblock \bibinfo{title}{{S}em{E}val-2025 {T}ask 8: {Q}uestion {A}nswering over {T}abular {D}ata},
\newblock in: \bibinfo{booktitle}{Proceedings of the 19th International Workshop on Semantic Evaluation (SemEval-2025)}, \bibinfo{publisher}{Association for Computational Linguistics}, \bibinfo{address}{Vienna, Austria}, \bibinfo{year}{2025}.
\bibitem[{Zhou et~al.(2022)Zhou, Hu, Dong, Cheng, Han, and Zhang}]{zhou2022tacubeprecomputingdatacubes}
\bibinfo{author}{F.~Zhou}, \bibinfo{author}{M.~Hu}, \bibinfo{author}{H.~Dong}, \bibinfo{author}{Z.~Cheng}, \bibinfo{author}{S.~Han}, \bibinfo{author}{D.~Zhang}, \bibinfo{title}{Tacube: Pre-computing data cubes for answering numerical-reasoning questions over tabular data}, \bibinfo{year}{2022}. \URLprefix \url{https://arxiv.org/abs/2205.12682}. \href{http://arxiv.org/abs/2205.12682}{{\tt arXiv:2205.12682}}.
\bibitem[{He(2024)}]{he2024achievingtoolcallingfunctionality}
\bibinfo{author}{S.~He}, \bibinfo{title}{Achieving tool calling functionality in llms using only prompt engineering without fine-tuning}, \bibinfo{year}{2024}. \URLprefix \url{https://arxiv.org/abs/2407.04997}. \href{http://arxiv.org/abs/2407.04997}{{\tt arXiv:2407.04997}}.
\bibitem[{Sheu et~al.(2005)Sheu, Chang, and Huang}]{1423558}
\bibinfo{author}{S.~Sheu}, \bibinfo{author}{A.~Chang}, \bibinfo{author}{W.~Huang},
\newblock \bibinfo{title}{Fast similarity search in string databases},
\newblock in: \bibinfo{booktitle}{19th International Conference on Advanced Information Networking and Applications (AINA'05) Volume 1 (AINA papers)}, volume~\bibinfo{volume}{1}, \bibinfo{year}{2005}, pp. \bibinfo{pages}{617--622 vol.1}. \DOIprefix\doi{10.1109/AINA.2005.185}.

\end{thebibliography}

\newpage
\appendix
\section{Appendix I: Custom functions for coder}
\label{AppendixI}

This section contains the definitions of the functions that have been implemented to help the coder to perform some common operations.

\begin{lstlisting}
def flatten_column_values_from_df(df: pd.DataFrame, column: str) -> pd.DataFrame:

def get_top_n_records_with_non_nan_column_value(
df: pd.DataFrame, column: str, number: int
) -> pd.DataFrame:

def get_tail_n_records_with_non_nan_column_value(
df: pd.DataFrame, column: str, number: int
) -> pd.DataFrame:

def delete_rows_by_column_value(
df: pd.DataFrame, column: str, value: Any = None
) -> pd.DataFrame:

def sort_dataframe_column_alphabetical_order(df: pd.DataFrame, column_name: str):

def filter_rows_by_column_equals_or_less_than_numeric_value(
df: pd.DataFrame, column: str, value: Any = None
) -> pd.DataFrame:

def filter_rows_by_column_strictly_less_than_numeric_value(
df: pd.DataFrame, column: str, value: Any = None
) -> pd.DataFrame:

def filter_rows_by_column_equals_or_higher_than_numeric_value(
df: pd.DataFrame, column: str, value: Any = None
) -> pd.DataFrame:

def filter_rows_by_column_strictly_higher_than_numeric_value(
df: pd.DataFrame, column: str, value: Any = None
) -> pd.DataFrame:

def filter_rows_that_contain_column_value(
df: pd.DataFrame, column: str, value: Any = None
) -> pd.DataFrame:

def filter_rows_that_do_not_contain_column_value(
df: pd.DataFrame, column: str, value: Any = None
) -> pd.DataFrame:

def exists_value_in_column(df: pd.DataFrame, column: str, value) -> bool:

def count_elements_equal_to_value_in_column(
df: pd.DataFrame, column: str, value: Any = None
) -> int:

def count_elements_containing_value_in_column(
df: pd.DataFrame, column: str, value: Any = None
) -> int:

def find_n_most_frequent_elements_in_column_subset(
df: pd.DataFrame, target_column: str, subset_column: str, filter_value, n: int
) -> list:

def find_most_frequent_element_in_column_subset(
df: pd.DataFrame, target_column: str, subset_column: str, filter_condition
):

def find_most_frequent_element_in_column(df: pd.DataFrame, column: str = None):

def find_n_most_frequent_elements_in_column(
df: pd.DataFrame, column: str, n: int
) -> list:
\end{lstlisting}

\section{Appendix II: Fuzzy filters}
\label{AppendixII}

This section contains the code of the functions that implement the fuzzy match filtering in order to explain the steps that this filtering follows.

\begin{lstlisting}
def _best_fuzzy_match(
    series: pd.Series, target: str, threshold: int = 90
) -> Any | None:
    unique_vals = series.dropna().unique()
    best_val, best_score = None, 0
    for v in unique_vals:
        score = fuzz.ratio(str(v).lower(), target.lower())
    if score > best_score:
        best_val, best_score = v, score
        return best_val if best_score >= threshold else None

def filter_rows_that_contain_column_value(
    df: pd.DataFrame, column: str, value: Any = None
) -> pd.DataFrame:
    threshold = 75
    # ---- round 1: exact / simple contains ---------------------------------
    # ---- round 2: fuzzy ----------------------------------------------------
    if (
        pd.api.types.is_string_dtype(df[column])
        and isinstance(value, str)
        and value != ""
    ):
        best_match = _best_fuzzy_match(df[column], value, threshold)
        if best_match is not None:
            fuzzy = df[df[column] == best_match]
        if _round_was_useful(original_len, len(fuzzy)):
            return fuzzy
\end{lstlisting}


\end{document}